\documentclass[conference]{IEEEtran}
\usepackage{cite}
\usepackage{amsmath,amssymb,amsfonts}
\usepackage[]{algorithm2e}
\usepackage{graphicx}
\usepackage{textcomp}
\usepackage{xcolor}
\usepackage{pifont}
\usepackage{caption}
\usepackage{changepage}
\usepackage{hyperref}
\usepackage{array}
\newcolumntype{?}{!{\vrule width 1pt}}

\def\BibTeX{{\rm B\kern-.05em{\sc i\kern-.025em b}\kern-.08em
    T\kern-.1667em\lower.7ex\hbox{E}\kern-.125emX}}

\newcommand{\overbar}[1]{\mkern 1.5mu\overline{\mkern-1.5mu#1\mkern-1.5mu}\mkern 1.5mu}

\begin{document}

\title{Temporal Action Localization Using Gated Recurrent Units\\
}

\author{
\IEEEauthorblockN{Hassan Keshvarikhojasteh}
\IEEEauthorblockA{\textit{Sharif University of Technology} \\
Tehran, Iran \\
ha.keshvari@student.sharif.edu}
\and
\IEEEauthorblockN{Hoda Mohammadzade}
\IEEEauthorblockA{\textit{Sharif University of Technology} \\
Tehran, Iran \\
hoda@sharif.edu}
\and
\IEEEauthorblockN{Hamid Behroozi}
\IEEEauthorblockA{\textit{Sharif University of Technology} \\
Tehran, Iran \\
behroozi@sharif.edu}
}

\maketitle

\begin{abstract}
Temporal Action Localization (TAL) task which is to predict the start and end of each action in a video along with the class label of the action has numerous applications in the real world. But due to the complexity of this task, acceptable accuracy rates have not been achieved yet, whereas this is not the case regarding the action recognition task. In this paper, we propose a new network based on Gated Recurrent Unit (GRU) and two novel post-processing methods for TAL task. Specifically, we propose a new design for the output layer of the conventionally GRU resulting in the so-called GRU-Split network. Moreover, linear interpolation is used to generate the action proposals with precise start and end times. Finally, to rank the generated proposals appropriately, we use a Learn to Rank (LTR) approach. We evaluated the performance of the proposed method on Thumos14 and ActivityNet-1.3 datasets. Results show the superiority of the performance of the proposed method compared to state-of-the-art. Specifically in the mean Average Precision (mAP) metric at Intersection over Union (IoU) of 0.7 on Thumos14, we get 27.52\% accuracy which is 5.12\% better than that of state-of-the-art methods. 
\end{abstract}


\section{Introduction}

With ever-increasing number of videos available, especially alongside widespread use of internet and social media, video understanding has become much needed in a wide variety of applications including video summarization, human-machine interaction, visually impaired individuals. Without using an automated machine, manual extraction of required information from very large number of videos takes much time and costs much money. TAL is one of the most important topics in this area, in which the objective is to predict temporal boundaries and labels of various actions in a video.
 
Different methods have been proposed for the TAL task in recent years \cite{ssad}- \cite{tsa}, which can be divided into two categories: 1. One-stage approaches 2. Two-stage approaches. In one-stage approaches, the start and end times of each action and its label are generated in one step, while in two-stage approaches, first, the proposals with high recall are generated, and then they are classified to predict their labels. Single Shot Temporal Action Detection (SSAD) \cite{ssad} is an example of one-stage methods, which skips the proposal generation step and directly produces the label, start and end times, and the confidence score of each anchor. SSAD generates anchors with multiple scales for each cell of the extracted feature map. Boundary Sensitive Network (BSN) \cite{bsn}, as an example of two-stage approaches, tries to solve the previous method’s drawbacks and produce proposals with precise temporal boundaries, flexible length, and reliable confidence score. BSN combines high probability temporal boundaries to generate proposals. It then retrieves the proposals containing actions within them. 

Despite the improvement made by BSN compared to state-of-the-art methods, it cannot handle videos with various lengths, and as a result, its performance deteriorates under such circumstances. To alleviate this issue, RecapNet \cite{recapnet} which was inspired by human video understanding was proposed. Authors introduce a novel residual causal convolution module to keep and use past information. This network can deal with various length videos. 

\begin{figure*}[!t]
	\centering
	\includegraphics[width= \linewidth]{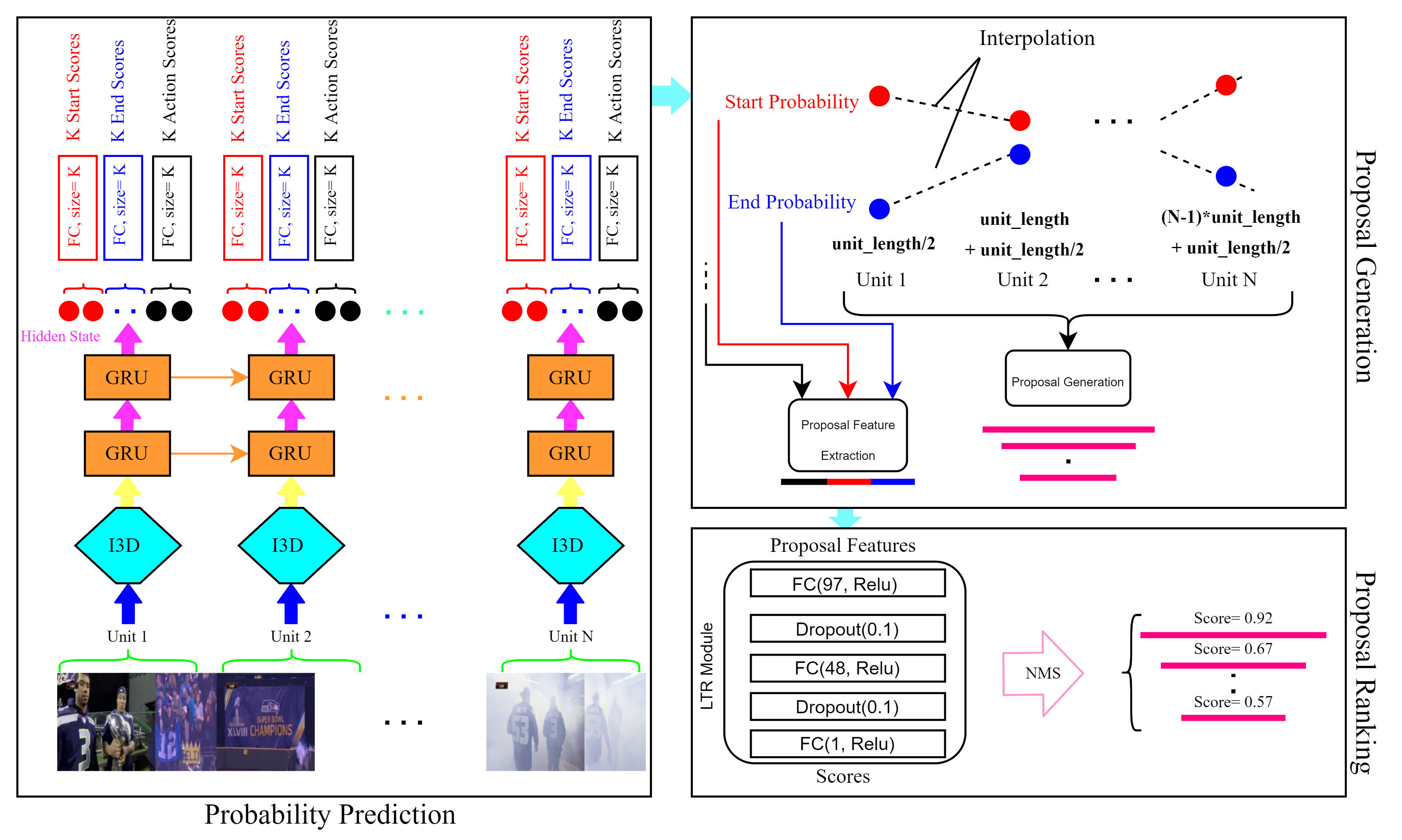}
	\caption{Our proposed method. In the first stage, using the GRU-Split network, we predict the start, end, and action probabilities
		for each input unit. Then, we generate candidate proposals using these probabilitits and predict their temporal boundaries by interpolation. Also, we extract each proposal features in this stage. In the final stage, proposals are ranked and selected by LTR module and NMS, respectively.}
	\label{fig: model}
\end{figure*}

In this paper, we propose the GRU-Split network in which the hidden states of each GRU cell are divided equally into three parts. One of these parts is used to predict the start probability, one part is used to predict the end probability, and the other part is used to predict the action probability. Similar to the convention of previous methods \cite{bsn, recapnet}, each video is divided into many consecutive non-overlapping parts, where each part consists of a fixed number of frames and is called a unit. Using the GRU-Split network, the proposals are generated in terms of the units and therefore, their temporal boundaries are required to be transformed into the time domain. State-of-the-art methods choose the middle frame of the start unit as the start frame and likewise the end frame is selected. In order to find precise temporal boundaries in the time domain, we use a linear interpolation approach to obtain the start and end probabilities for all the frames within each. This interpolation method not only enhances the performance of GRU-Split, but also the performance of GRU and RecapNet models \cite{recapnet}. Using GRU-Split network and the proposed interpolation method, we can obtain proposals with high recall and precise boundaries. Thereafter, the generated proposals are required to be ranked. Recent methods like BSN and TSA-Net \cite{tsa} compute the score of each proposal by using a fully connected network, where the inputs of this network are the predicted probabilities. To train the network, in each iteration, a batch of proposals are chosen randomly from different videos and the loss is calculated using a predefined score for each proposal. The problem with this method is that each proposal is treated independently and so the relationship between the proposals are ignored. In order to alleviate this issue, we feed the proposals of each video separately to a network designed by a Learn to Rank (LTR) approach, which improves the results. Our proposed network is depicted in \figurename{ \ref{fig: model}}. In summary, our contributions are as follows:

\begin{enumerate}
	\setlength{\itemsep} {0pt}
	\setlength{\parskip} {0pt}
	\item Introducing GRU-Split network to generate the start, end and action probabilities efficiently.
	\item Generating the proposals with precise temporal boundaries using a linear interpolation method. 
	\item Ranking the generated proposals effectively with an LTR approach. 
	\item Achieving better results compared to the state-of-the-art methods based on three different metrics on Thumos14 dataset, which is known as a golden dataset for the TAL task.
\end{enumerate}

\section{Related Works}

Since action recognition is a fundamental task in video understanding, we first briefly review action recognition methods and then go through the TAL and LTR methods. 
\subsection{Action recognition} 
Due to its importance, action recognition has received a lot of attention in the literature. Conventional methods are based on extracting hand-crafted features, such as spatio-temporal interest points \cite{eventdet}, improved dense trajectory (iDT) \cite{idt}, etc. With the advent of deep learning techniques, the task of extracting the features has been transferred to deep networks. As a result of extracting specialized features from video sequences using complex network architectures, impressive results in action recognition have been achieved over the past years \cite{twostream}– \cite{latetemp}. Deep learning methods in action recognition can be categorized into three main groups: methods based on Convolutional Neural Network (CNN), methods based on Recurrent Neural Network (RNN) and other architectures. Various extensions of CNN such as Pose-based CNN \cite{pcnn}, 3D CNN \cite{threedtpami}, Long Term Temporal Convolution \cite{longterm}, and I3D network \cite{quo} are examples of the first category. Many extensions of Long Short-Term Memory (LSTM) network such as global context-aware attention LSTM \cite{globalcontext} and spatio-temporal LSTM \cite{spationlstm} are examples of the second category. Late Temporal Modeling with BERT \cite{latetemp} is an example of the third category, in which a Bidirectional Encoder Representations from Transformers (BERT) layer \cite{bert} is used on top of a 3D CNN to capture temporal information using attention technique. This method has achieved 98.69\% top-1 accuracy on the UCF-101 \cite{ucf101} dataset, which is an outstanding result for the action recognition task. 

\subsection{Temporal Action Localization}

\begin{figure*}[!t]
	\centering
	\includegraphics[width= \linewidth]{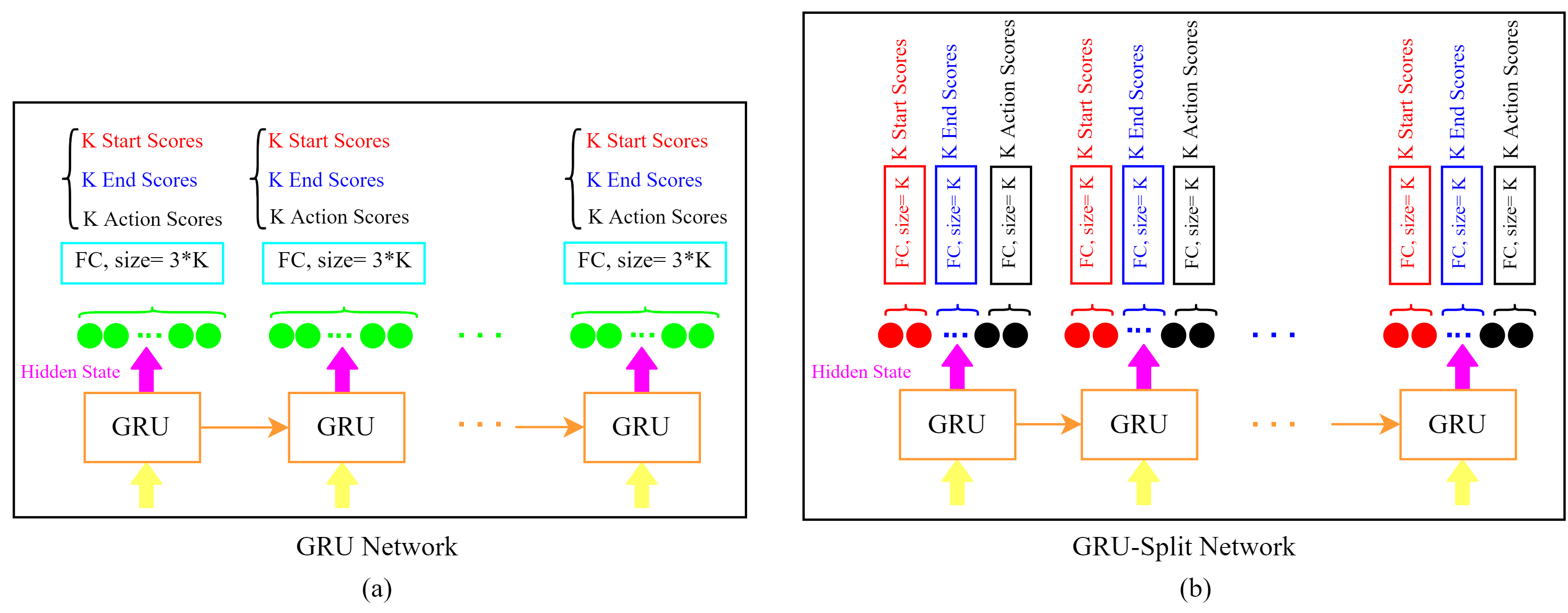}
	\caption{GRU network vs. GRU-Split network. (a). GRU network uses one FC layer to predict the probabilities, (b). For the GRU-Split network, we split the hidden state in each step to predict the probabilities separately.}
	\label{fig: gru-splitted}
\end{figure*}

As mentioned in Introduction section, TAL methods can be categorized into one-stage and two-stage methods. G-TAD (Sub-Graph Localization for Action Detection) \cite{gtad} is an example from the first category. In this method, a video is represented by a graph, where each unit of the video is a node, the temporal relationships between the units constitute the temporal edges and the semantic features of the units form the semantic edges of the graph. Four types of nodes are defined in the graph: start, end, action, and background.  TAL is then cast as a subgraph localization problem which is performed using the proposed Graph Convolutional Network (GCN) to incorporate semantic and temporal properties of the nodes simultaneously.  In \cite{relaattention}, it is argued that recognizing generated proposals individually is difficult due to the large within-class variation of actions. On the other hand, different proposals in a video may share important information regarding an action. Hence, in order to take into account the relation between the proposals, relation attention module is proposed based on the self-attention mechanism, which has been successful in capturing dependency between words in automatic translation. In \cite{graphconv}, a GCN is used to model the relation between the proposals. In order to construct the graph, the action proposals are represented by the nodes and two types of edges are introduced: contextual and surrounding edges. If two proposals have temporal overlap between them, they are connected by contextual edges and otherwise they are connected by surrounding edges, if the temporal distance between their centers is less than a threshold. Graph convolutional layers are then implemented using the adjacency matrix and ReLU activation function is used after each layer. The method proposed in \cite{progbrefinement} benefits from both anchor-based and frame-based approaches by using three cascaded detection modules. The first module detects coarse boundaries using pyramidal features from anchors. The second module uses pyramidal features to refine the anchor-based detections of the first module, and the third module refines the resulting boundaries using frame-based features.  In \cite{salad}, the network consists of a 3D CNN as the backbone and a bidirectional GRU to generate latent vectors in order to regress the start, end ,and class probabilities. The main contribution of this work is the use of self-assessment which allows to prune frames based on their IoU and score during training. In \cite{polo}, both RGB and flow features are used for TAL. But instead of early or late fusion of these features, a cross-modality fusion is proposed in order to incorporate there features effectively. In the proposed fusion framework, frame-level attention weights are learned from RGB features and employed to the flow features and vice versa. In \cite{revisiting}, an anchor-free TAL method is proposed in which an action instance is represented as a point with its distances to the start and end boundaries. Using this method, action instances with extremely short or long durations can be better predicted compared to anchor-based methods. As a result, better overall predictions are achieved by combining the proposed anchor-free method with an anchor-based approach. Another weakly supervised approach is followed in \cite{knowdist}. In this method, by using a knowledge distillation framework, snippet-level pseudo labels are generated first. To generate these labels reliably, cross-modality knowledge is distilled by transferring knowledge between RGB and flow streams and the fused-modal stream. The noise of the generated pseudo labels are then reduced by using a multi-modal self-voting scheme.

TSA-Net\cite{tsa} is an example of two-stage methods. This model uses an ensemble of dilated convolution filters to expand the receptive field for considering actions with different temporal durations which is ubiquitous in TAL datasets. Multi Dilation Temporal Convolution (MDC) block is designed, and multiple branches consisting of stacked MDC blocks are used to predict probabilities. In \cite{subaction}, a weakly supervised method is proposed to learn the start, end, and category label of untrimmed videos using video-level category labels of training videos. In this method, a sub-action family representation, which is inspired by the well-known bag-of-words object classification method, is first constructed using the video-level category labels of training videos. The features extracted from the videos are then compared with the sub-actions and the resulting scores are used to infer the temporal boundaries of the action instances.
As an interesting research direction, some researchers benefit from reinforcement learning for TAL task. In \cite{sap}, a self-adapted model called Self-Adaptive Proposal is proposed. In this model, TAL is cast as a Markov decision process, in which the agent observes the current video segment and adjusts the action proposals to obtain precise proposals. After each decision, the agent receives adequate rewards to learn a proper policy.  
 
\subsection{Learn To Rank}
\label{subsec: ltr}
In most TAL methods, first a large number of overlapping proposals are generated and then they are ranked to select the proposal with the highest score. An appropriate ranking has a significant impact on TAL performance, as shown in \cite{bottomup}. To perform the ranking, in \cite{recapnet, fastlearning} predefined scores, which are functions of start, end, and action probabilities, are used to predict the score of the proposals, but the performance of these methods is not satisfactory. In \cite{bsn} and \cite{tsa}, deep learning networks are used to learn the score of the proposals using their overlap with the ground-truth regions. In these methods, the typical architecture is a (Fully Connected) FC network in which the input is the so-called boundary sensitive proposal features, generated using the predicted probabilities, and the output is the predicted scores. This network is called Rank Module and the final score for each proposal [$u_{i}, u_{j}$] is $S(ij) = P_{s_i} * P_{e_j} * \varphi(ij)$ where $P_{s_i}$, $P_{e_i}$, and $\varphi(ij)$ are start probability, end probability, and the score predicted by network, respectively.

 In information retrieval, Learn to Rank (LTR) is used to sort documents based on their relevance to a target word. The goal of LTR is to find a model to rank objects appropriately, which is required in different applications like document retrieval, product rating, etc. Three main approaches exist for LTR. In the first approach, which is called pointwise, the problem is transformed to a regression or classification on a single object \cite{discmodels}. The pairwise approach is the second one that tries to learn differences between pairs of objects \cite{ltr, svm}. These two approaches ignore the relationships between the objects as a list. The listwise approach alleviates the drawbacks of the previous approaches by taking list of objects as samples in learning. In \cite{bpairwise2listwise}, in order to define a loss function for lists of objects, two probabilities are proposed namely permutation and top one probabilities. A neural network, called ListNet, is then proposed for learning, which uses cross-entropy between normalized predicted and ground truth scores based on top one probability as metric. The proposed method is then used to rank the list of documents for information retrieval. To the best of our knowledge, for the first time, we use a similar method to rank our generated action proposals.  

\section{Our Approach}
Given an untrimmed video, consisting of instances of various actions with different temporal durations, we need to generate proposals with true labels and large overlap with ground truth. In the following, different parts of our proposed method are described.
\subsection{Video feature encoding}
\label{subsec: vfe}
As mentioned in Introduction, we divide a video into units, which are consecutive non-overlapping parts, consisting of a fixed number of frames. We extract the spatial and temporal features of each unit using pretrained two-stream I3D network \cite{quo}, as shown on the left of \figurename{ \ref{fig: model}}. Extracting features from units instead of frames, significantly reduces the computational cost of TAL. The two-stream I3D network is fed with stacked RGB frames to extract spatial features, and with stacked optical flows to extract temporal features. By concatenating the resulting features, the final feature vectors of 2048 dimensions are obtained. 

\subsection{GRU-Split} 

The extracted features are fed to our GRU-Split network for predicting the start, end, and action probabilities, as shown in the left side of \figurename{ \ref{fig: model}}. We use the same approach as RecapNet \cite{recapnet} to train our network, i.e., in each step, the network predicts these probabilities for the current unit and past K-1 units. It is important to note that although only the features of the current unit are fed to the GRU cell at each step, but using the information stored in the hidden states, the network can predict the probabilities for the previous K-1 units. For the sake of comparison, the architecture of the conventional GRU network is shown in \figurename{ \ref{fig: gru-splitted}}(a). The last layer of this network is a Fully Connected (FC) layer for transforming the hidden states of each GRU cell into probabilities. In contrast, our proposed GRU-Split network, uses three FC layers with separated inputs as shown in \figurename{ \ref{fig: gru-splitted}}(b). Each FC layer predicts one of the three probabilities (start, end, and action) so each of them can learn to predict the corresponding probability more accurately compared to the previous architecture.  By splitting the hidden states, the error of each output does not affect the parameters of the hidden state and FC layer of the other two outputs and therefore, the parameters related to each other become more specialized in estimating that output. In other words, for example, the start probability might be more erroneous compared to the action probability and we do not want any of them to get attenuated in favor of the other one. In Experiment Section, we discuss the performance and size of the GRU-Split model.

\subsection{Proposal Generation} 
The next step after predicting probabilities is the proposal generation. Up to this point, for each unit, we have predicted K start probabilities, K end probabilities, and K action probabilities. In order to generate the action proposals, we use a method similar to RecapNet \cite{recapnet}, i.e., start and end units are selected using two rules: voting scheme, and peak value picking. In the voting scheme, units having more than V start/end probabilities greater than a threshold, are chosen as start/end candidates. On the other hand, in the peak value picking procedure, units with average start/end probability greater than their adjacent units (one past and one next) are chosen. Both of these rules are used simultaneously to detect candidate start and end units. A start unit matches an end one if the former is preceded by the latter. 

The scores of these proposals can be computed using the following function as similar to \cite{recapnet}:

\begin{equation}
S(ij)= \overbar{P_{s_i}}*\overbar{P_{e_j}}*\dfrac{\sum_{k= i}^{j}\overbar{P_{a_k}}}{|j-i|} 
\label{eq:1}
\end{equation}
where $\overbar{P_{s_i}}$ is the average of K start probabilities of unit $i$,  $\overbar{P_{e_j}}$ is the average of K end probabilities of unit $j$, and $\overbar{P_{a_k}}$ is the average of K action probabilities of unit $k$.

\subsection{Ranking Module}
\label{subsec: RM}

\begin{figure}[!t]
	\centering
	\includegraphics[width=0.55\linewidth]{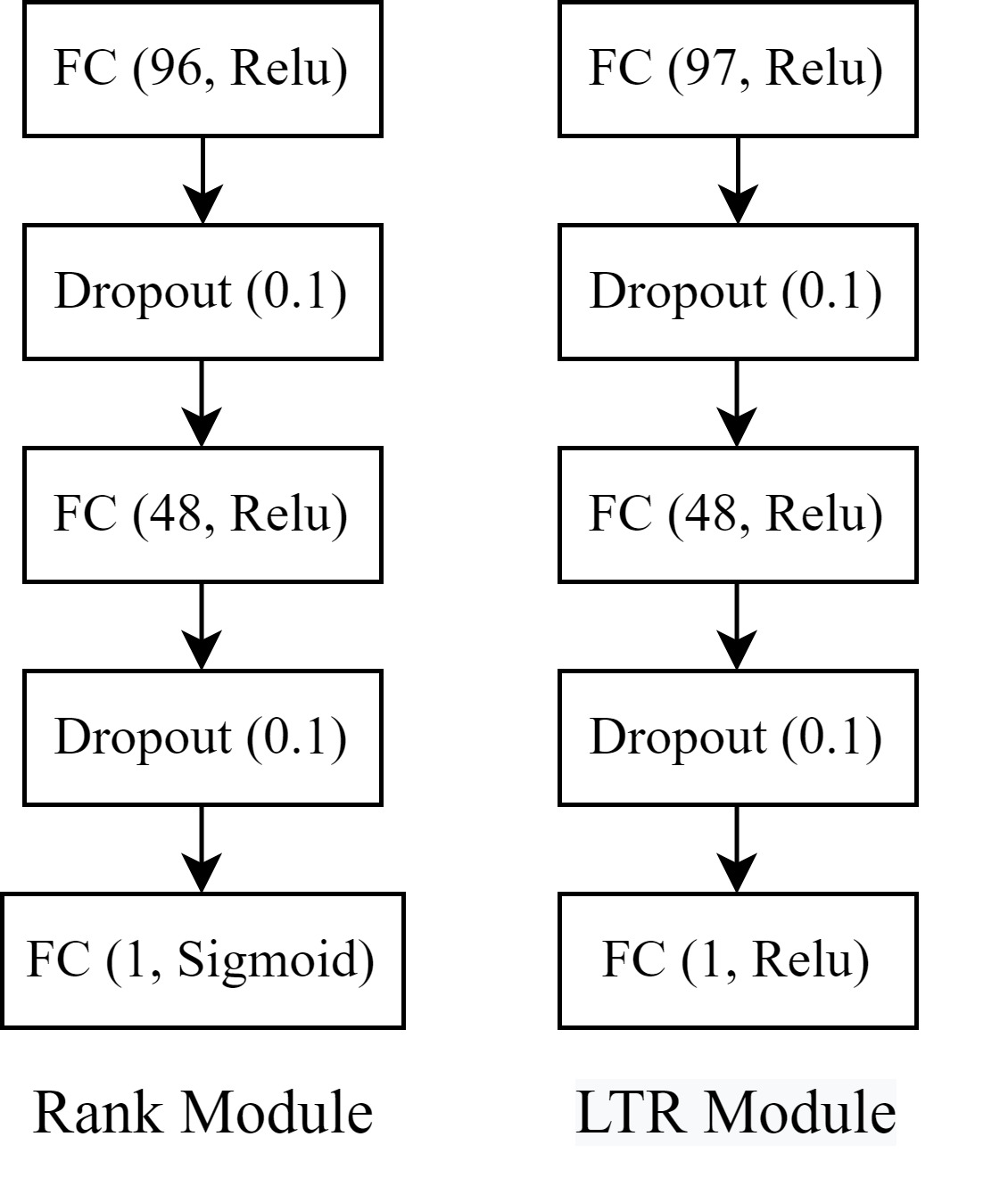}
	\caption{Ranking modules.}
	\label{fig: rm}
\end{figure}

Thus far, we have assigned to each unit, one start, one end,    and one action probability set followed by generating action proposals with high recall rate. As many overlapping proposals are generated for each video, ranking them is necessary at this point. As mentioned in Introduction section, we use LTR \cite{bpairwise2listwise}, which is a listwise ranking method for documents, to rank the generated proposals. In this method, features of all document in a list are extracted and concatenated with the features of the query word, and fed to a network for ranking. Using these features as input, the network learns the relevance of each document to query. We use a method similar to LTR to rank the generated proposals. In our ranking problem, we have a list of proposals for each video. In our proposed ranking method, the video plays the role of query word, and the generated proposals play the role of the list of documents. For each proposal, we construct a 96-dimensional feature vector using its start, end and action probabilities as follows. Let’s [$u_s$,$u_e$] indicate the region of a candidate proposal, where $u_s$ denotes the start unit and $u_e$ denotes the end unit. The proposal region is divided into three regions: start, end, and action region. The action region is defined as [$u_s$,$u_e$], the start region as [$u_s-d/5$,$u_s+d/5$], and the end region as [$u_e-d/5$,$u_e+d/5$], where $d=u_e-u_s$. We then sample 16 action probabilities uniformly from the action region by linear interpolation. Also, we sample 8 action probabilities from the start and end regions resulting in total of 32 sampled action probabilities. Similarly, we sample 32 start and end probabilities from these regions. The resulting 96 probabilities for each proposal, construct its feature vector. It should be noted that our method for extracting proposal features is similar to the method used in \cite{bsn}, but instead of sampling only action probabilities, we sample from all three probabilities which results in more discriminative features for proposals. As the feature of the video, we concatenate with the features of every proposal, the length of the video normalized to maximum video length in the dataset. The resulting 97 features are fed to our LTR module to predict the scores of each proposal. We want that the predicted scores show the overlap between the predicted proposals and the ground truth action regions. The overlap for the $ith$ proposal is defined as:

\begin{figure}[!t]
	\centering
	\includegraphics[width=\linewidth]{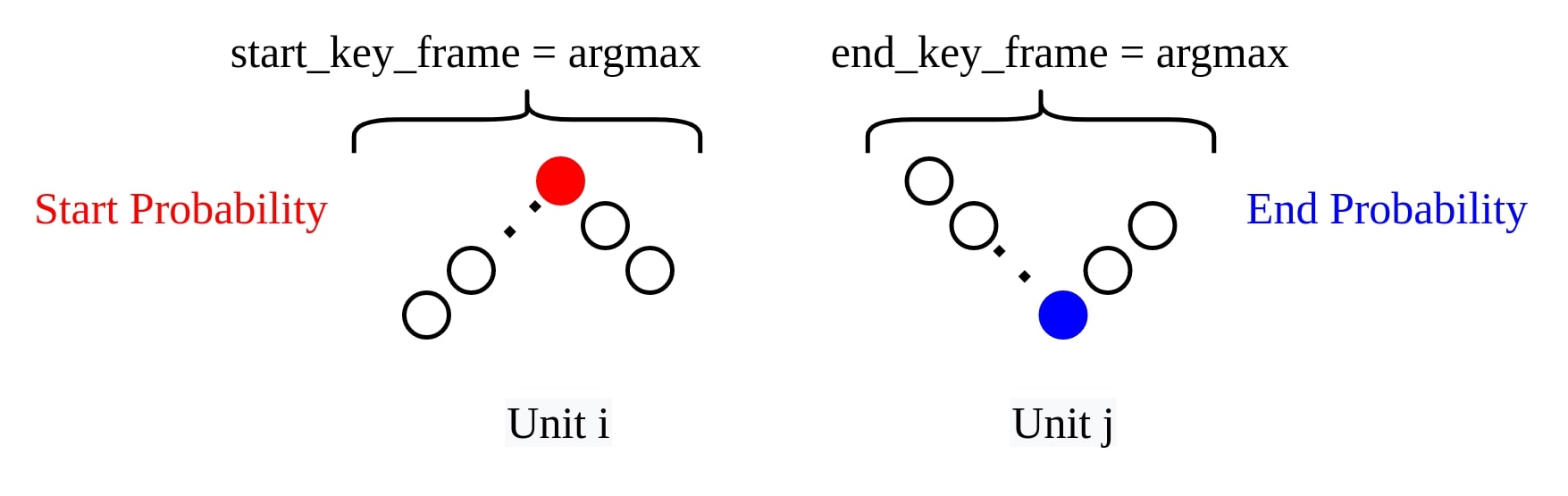}
	\caption{Using interpolated probabilities to find the start and end keyframes. The colored circles represent the probabilities of units  and the empty circles are interpolated probabilities.}
	\label{fig: interpolation}
\end{figure}

\begin{table*}[!t]
	\centering
	\caption{GRU-Split configuratin}
	\renewcommand{\arraystretch}{1.5}
	\begin{tabular}{|c|c|c|c|c|c|c|c|}
		\hline
		\textbf{Layers} & \textbf{Hidden-states} & \textbf{Feat-dims} & \textbf{K}&\textbf{Dropout} & \textbf{L2-norm } & \textbf{Loss function} & \textbf{Optimizer}\\ 
		\hline
		2 & 513 & 2048 & 17 &0.2 & 1e-5 & Cross-Entropy & Adam\\
		\hline
	\end{tabular}
	\label{tab: parameters}
\end{table*}

\begin{table*}[!t]
	\centering
	\caption{GRU network performance using different parameters on the validation set based on AR@AN and R@AN=100-tIoU metrics on Thumos14}
	\renewcommand{\arraystretch}{1.5}
	\begin{tabular}{|c?c|c|c|c|c?c|c|c|c|c?}
		\hline
		\textbf{Parameters}& \multicolumn{5}{c?}{\textbf{AR@AN}} & \multicolumn{5}{c?}{\textbf{R@AN=100-tIoU}}\\
		\cline{2-11}
		&\textbf{@50} & \textbf{@100} & \textbf{@200} & \textbf{@300} & \textbf{@400} & \textbf{0.5} & \textbf{0.6} & \textbf{0.7} & \textbf{0.8} & \textbf{0.9}\\ 
		\hline
		V= 3, thresh= 0.5&33.63 & 40.78 & 47.87 & 50.72 & -&69.26 & 62.38 & 54.12 & 42.20 & 14.06\\
		\hline
		V= 3, thresh= 0.3&34.67 & 42.20 & 50.44 & 55.26 & 57.25&\textbf{71.25} & 64.22 & 55.96 & 44.64 & 14.67\\ 
		\hline 
		V= 1, thresh= 0.5&34.60 & 42.71 & 50.10 & 54.52 & 56.85&70.79 & 64.52 & 56.42 & 45.56 & 15.44\\ 
		\hline
		\textbf{V= 1, thresh= 0.3}&\textbf{35.22} & \textbf{42.97} & \textbf{51.11} & \textbf{55.32} & \textbf{58.03}& 70.94 & \textbf{64.52} & \textbf{56.42} & \textbf{46.48} & \textbf{16.05}\\ 
		\hline
	\end{tabular}
	\label{tab: parameter tunning}
\end{table*}

\begin{equation}
g(i) = \underset{j = 1, ...}{max} IoU\{P(i), G(j)\}
\end{equation}
where $G(j)$ denotes the $jth$ ground-truth action region. In order to train the LTR module with appropriate scores, we relate the proposal scores to each other using the softmax function:
\begin{equation}
\hat{g}(i) = \dfrac{\exp \tilde{g}(i)}{\underset{_{j = 1, ...}}{\sum} \exp \tilde{g}(j)}
\end{equation}

where: 
\[\tilde{g}(i) = 
\begin{cases}
	0, & 0 <= g(i) <= 0.2\\
	1, & 0.2 < g(i) <= 0.4\\
	2, & 0.4 < g(i) <= 0.6\\
	3, & 0.6 < g(i) <= 0.8\\
	4, & 0.8 < g(i) <= 0.9\\
	5, & 0.9 < g(i) <= 1
\end{cases}
\]

We feed the proposals of each video in each iteration to our network to predict the modified overlap score of each proposal ($\varphi(i)$). The architecture of our LTR module is shown in \figurename{ \ref{fig: rm}}. Cross-Entropy between normalized predicted scores and ground truth scores is used as the loss function:

\begin{equation}
	L_{LTR-M} = \dfrac{1}{N} \sum_{i=1}^{N} -\hat{g}(i)log\hat{\varphi}(i)
\end{equation}
where:
\[
\hat{\varphi}(i) = \dfrac{\exp \varphi(i)}{\underset{_{j = 1, ...}}{\sum} \exp \varphi(j)}
\] 

%
As a result of using this method, unlike the rank modules used in previous works \cite{bsn, tsa}, which predict the score of each proposal separately, LTR module takes into account the relationship between all proposals of each video and learns to rank them with respect to each other. 

\subsection{Frame Probability Interpolation}
After detecting the proposal in the unit domain, they should be transformed to the frame and then time domain. For such transformation, in RecapNet \cite{recapnet} and other state-of-the-art methods \cite{bsn, fastlearning}, the middle frame of the start/end units is considered as the start/end frame, and the proposal temporal boundaries are calculated using the frame per second (fps) rate of each video. For example, suppose the predicted proposal is [$ui$, $uj$], where $ui$ is the $ith$ unit and $uj$ is the $jth$ unit of the video. For the transformation to the time domain, the following equations are used:

\[t_i = \dfrac{u_i* (unit\underline{\hspace{5pt}}length) + (\dfrac{unit\underline{\hspace{5pt}}length}{2})}{fps}\]
\begin{equation}
	t_j = \dfrac{u_j* (unit\underline{\hspace{5pt}}length) + (\dfrac{unit\underline{\hspace{5pt}}length}{2})}{fps}
	\label{eq: time}
\end{equation}  

Because using the middle frame as the start/end frame is not accurate, we propose to use an interpolation approach to find these frames. As before, we assign the average of the K start/end probabilities of each unit to its middle frame, but then we interpolate the average probabilities over other frames. We then select the frame with the maximum start/end probability as the start/end frames. This method is shown in \figurename{ \ref{fig: interpolation}}. The start and end times are therefore obtained as:

\[t_i^{'} = \dfrac{u_i* (unit\underline{\hspace{5pt}}length) + (start\underline{\hspace{5pt}}key\underline{\hspace{5pt}}frame)}{fps}\]
\begin{equation}
	t_j^{'} = \dfrac{u_j* (unit\underline{\hspace{5pt}}length) + (end\underline{\hspace{5pt}}key\underline{\hspace{5pt}}frame)}{fps}
	\label{eq: time_ref}
\end{equation}

We should mention that as we did not see any more benefit with other interpolation methods, we chose linear interpolation.

\section{Experiments}

\begin{table*}[!t]
	\caption{Comparison with state-of-the-art models on the test set based on AR@AN, R@AN=100-tIoU, and mAP metrics on Thumos14}
	\centering
	\renewcommand{\arraystretch}{1.5}
	\begin{tabular}{|c?c|c|c|c|c?c|c|c|c|c?c|c|c|c|c?}
		\hline
		\textbf{Method} & \multicolumn{5}{c?}{\textbf{AR@AN}} &  \multicolumn{5}{c?}{\textbf{R@AN=100-tIoU}} & \multicolumn{5}{c?}{\textbf{mAP}}\\
		\cline{2-16}
		& \textbf{@50} & \textbf{@100} & \textbf{@200} & \textbf{@300} & \textbf{@400} & \textbf{0.5} & \textbf{0.6} & \textbf{0.7} & \textbf{0.8} & \textbf{0.9} & \textbf{0.3} & \textbf{0.4} & \textbf{0.5} & \textbf{0.6} & \textbf{0.7}\\
		\hline
		\textbf{SST\cite{sst}} & 19.9 & 28.36 & 37.90 & 44.27 & 48.75 & 52.13 & 44.78 & 37.98 & 28.20 & 6.60 & 41.2  & 31.5 & 20.0 & 10.9 & 4.7\\
		\hline
		\textbf{TURN\cite{ttap}} & 21.86 & 31.89 & 43.02 & 49.18 & 54.18 & 61.53 & 52.33 & 41.31 & 26.90 & 9.37 & 46.3 & 35.3 & 24.5 & 14.1 & 6.3\\
		\hline
		\textbf{TAL-Net\cite{Rercnn}} & 35.50 & 42.02 & 47.28 & 49.56 & 50.62 & - & - & - & - & - &  53.2 & 48.5 & 42.8 & 33.8 & 20.8\\
		\hline
		\textbf{CTAP\cite{ctap}} & 32.96 & 42.76 & 51.85 & 57.25 & 60.17 & 72.37 & 63.93 & 55.16 & 41.36 & 20.74 & - & -&29.9&-&-\\
		\hline
		\textbf{BSN\cite{bsn}} & 37.46 & 46.06 & 53.21 & 56.82 & 59.05 & 77.99 & 71.62 & 60.25 & 44.11 & 19.76 & 53.5 & 45.0 & 36.9 & 28.4 & 20.0\\
		\hline
		\textbf{BMN\cite{bmn}} & 39.36 & 47.72 & 54.70 & - & -  & - & - & - & - & - & 56.0 & 47.4 & 38.8 & 29.7 & 20.5\\
		\hline
		\textbf{MGG\cite{mgg}} & 39.93 & 47.75 & 54.65 & - & -  & - & - & - & - & - & 53.9 & 46.8 & 37.4 & 29.5 & 21.3\\
		\hline
		\textbf{TSA-Net\cite{tsa}} & 42.83 & 49.61 & 54.52 & - & - & - & - & - & - & - & 53.2 & 48.1 & 41.5 & 31.5 & 21.7\\
		\hline
		\textbf{RecapNet\cite{recapnet}} & 38.58 & 48.43 & 57.04 & 60.97 & 63.44 & 74.82 & 70.41 & 63.89 & 52.27 & 22.89& - & -&-&-&-\\
		\hline
		\textbf{ETP \cite{ptal}} & -&-&-&-&-&-&-&-&-&-&48.2 & 42.4 & 34.2 & 23.4 & 13.9 \\
		\hline
		\textbf{DBS \cite{vimprint}} & -&-&-&-&-&-&-&-&-&-& 50.6 & 43.1 & 34.3 & 24.4 & 14.7 \\
		\hline
		\textbf{FC-AGCN-P \cite{graphatt}} & -&-&-&-&-&-&-&-&-&-& 57.1 & 51.6 & 38.6 & 28.9 & 17.0 \\
		\hline
		\textbf{TSI \cite{tsi}} & 42.30&50.51& 57.24&-&-&-&-&-&-&-& \textbf{61.0} & \textbf{52.1} & 42.6 & 33.2 & 22.4 \\
		\hline
		\textbf{GRU-Split} & \textbf{45.53} & \textbf{52.64} & \textbf{59.05} & \textbf{62.32} & \textbf{64.13} & \textbf{81.91} & \textbf{77.65} & \textbf{69.48} & \textbf{56.14} & \textbf{24.69} & 54.62 & 50.22 & \textbf{44.40} & \textbf{36.40} & \textbf{27.52} \\
		\textbf{+ Interpolation} & & & & & & & & & & &&&&&\\
		\textbf{+ LTR Module} &  &  &  &  & & & & & &&&&&&\\
		\hline
	\end{tabular}
	\label{tab: rm-compar}
	\vspace{5pt}
\end{table*}

\begin{table}[!t]
	\centering
	\caption{Comparison with state-of-the-art models on the validation set based on AR@AN=100 and AUC metrics on ActivityNet-1.3}
	\renewcommand{\arraystretch}{1.5}
	\begin{tabular}{|c|c|c|}
		\hline
		\textbf{Method} & \textbf{AR@AN= 100 (Val)} & \textbf{AUC (Val)} \\
		\hline
		\textbf{TURN\cite{ttap}} & 49.73 & 54.16 \\
		\hline
		\textbf{Zhao et. al.\cite{zhao}} & 63.53 & 53.02\\
		\hline
		\textbf{TAG \cite{tag}} & 63.52 & 53.02 \\
		\hline
		\textbf{MSRA \cite{msra}} & - & 63.12 \\
		\hline
		\textbf{SSAD \cite{ssad}} & 73.01 & 64.40\\
		\hline
		\textbf{CTAP\cite{ctap}} & \textbf{73.17} & \textbf{65.72}\\
		\hline
		\textbf{Ours} & 72.43 & 63.92\\
		\hline
	\end{tabular}
	\label{tab: activity}
\end{table}

\subsection{Dataset}

\textbf{Thumos14} \cite{thumos} dataset is used for TAL as a benchmark dataset which consists of 200 untrimmed videos in the validation set and 213 untrimmed videos in the test set with temporal annotations of 20 classes. Train set for this dataset is UCF-101 dataset which consists only of trimmed videos and mostly used for action recognition task. Due to this reason, we only use the validation set to train the network and tune its parameters using 5-fold cross-validation on this set.

\textbf{ActivityNet-1.3} \cite{activity} dataset contains 19994 videos of 200 action classes in which the train set, validation set, and test set include 10024, 4926, and 5044 videos, respectively. The challenges for this dataset originate from the large number of action classes with varied scales.

\subsection{Metrics}

To evaluate the performance of the proposed method in proposal generation, AR@AN and R@AN = 100-tIoU are used. AR@AN computes average recall at different average number of proposals for each video with Intersection over Union (IoU) from 0.5 to 1 by stride 0.05 on Thumos14 and from 0.5 to 0.95 by stride 0.05 on ActivityNet-1.3. The R@AN = 100-tIoU metric computes recall at different IoU from 0.5 to 1 by stride 0.05 given the fixed average number of proposals (100) which is used on Thumos14. For ActivityNet-1.3, the area under the AR@AN curve (AUC) is computed as an another metric when AN varies from 0 to 100.  Also, mean Average Precision (mAP) at different IoUs from 0.3 to 0.7 by stride 0.1 is used to measure the ability of the method in action classification task on Thumos14.  

\subsection{Implementation Details}

\begin{table*}[!t]
	\centering
	\caption{Comparison between different networks performance on the test set based on AR@AN and R@AN=100-tIoU metrics on Thumos14}
	\renewcommand{\arraystretch}{1.5}
	\begin{tabular}{|c?c|c|c|c|c?c|c|c|c|c?}
		\hline
		\textbf{Method} &  \multicolumn{5}{c?}{\textbf{AR@AN}} & \multicolumn{5}{c?}{\textbf{R@AN=100-tIoU}}\\
		\cline{2-11}
		&\textbf{@50} & \textbf{@100} & \textbf{@200} & \textbf{@300} & \textbf{@400} & \textbf{0.5} & \textbf{0.6} & \textbf{0.7} & \textbf{0.8} & \textbf{0.9}\\ 
		\hline
		\textbf{GRU} & 39.64 & 47.21 & 54.43 & 58.72 & 61.25 & 75.24 & 70.32 & 62.69 & 49.38 & 20.51\\
		\hline
		\textbf{GRU-Split} & \textbf{41.55} & \textbf{49.51} & \textbf{57.15} & \textbf{61.07} & \textbf{63.41} & \textbf{79.99} & \textbf{74.64} & \textbf{65.57} & \textbf{50.57} & \textbf{20.93}\\ 
		\hline
	\end{tabular}
	\label{tab: gru-splitted}
\end{table*}	

\begin{table*}[!t]
	\caption{Comparison between different networks performance on the test set based on AR@AN and R@AN=100-tIoU metrics on Thumos14}
	\centering
	\renewcommand{\arraystretch}{1.5}
	\begin{tabular}{|c?c|c|c|c|c?c|c|c|c|c?}
		\hline
		\textbf{Method} &  \multicolumn{5}{c?}{\textbf{AR@AN}} & \multicolumn{5}{c?}{\textbf{R@AN=100-tIoU}}\\
		\cline{2-11}
		&\textbf{@50} & \textbf{@100} & \textbf{@200} & \textbf{@300} & \textbf{@400} & \textbf{0.5} & \textbf{0.6} & \textbf{0.7} & \textbf{0.8} & \textbf{0.9}\\ 
		\hline
		\textbf{GRU-Split} & 41.55 & 49.51 & 57.15 & 61.07 & 63.41 & 79.99 & 74.64 & 65.57 & 50.57 & 20.93\\
		\hline
		\textbf{GRU-Split} & \textbf{42.33} & \textbf{49.81} & \textbf{57.90} & \textbf{61.89} & \textbf{64.05}& \textbf{80.80} & \textbf{75.00} & \textbf{65.69} & \textbf{51.33} & \textbf{21.11}\\
		\textbf{+ Interpolation} & & & & &&&&&&\\
		\hline
	\end{tabular}
	\label{tab: frame}
\end{table*}	

As stated in section III-A, we use I3D model pretrained on the UCF-101 dataset to extract useful spatial and temporal features for TAL task. We use 16 consecutive frames without overlap as a unit to feed the I3D network. Also, for the training of the GRU-split network, we randomly select 100 consecutive units from each video to form a batch of features to be used as input to the network. We set the batch size to 32 for this network. We use two GRU-Split layers stacked over each other with 513 units as the hidden states and train the network for 200 epochs. The settings for this network such as, K, dropout, L2-norm, loss function, and optimizer are shown in \tablename{ \ref{tab: parameters}}, which are chosen similar to the GRU parameters in RecapNet \cite{recapnet}. 
%

For ActivityNet 1.3 dataset, the features are extracted using I3D network pretrained on Kinetics \cite{quo}. Sixteen consecutive frames without overlap are used as a unit and  the feature sequence of each video is rescaled to 100 length \cite{bsn, bmn}. Other settings are similar to those on Thumos14 dataset.

We tune the critical parameters like V and threshold of the GRU-split using the validation set. This is different from what is done in \cite{recapnet} to compare GRU and causal convolution, where the authors use the same parameters for two models and the superiority of causal convolution versus GRU is claimed. However, we know that parameter tuning in deep learning, like any other learning method is crucial and requires careful tuning for better results. As shown in \tablename{ \ref{tab: parameter tunning}}, V = 1 and threshold = 0.3 are the best parameters. The batch size of LTR module is varied, because we feed all proposals of each video in each iteration to it. We train the ranking modules for 20 epochs and use Adam as an optimizer with an initial learning rate of 1e-3 which decays by 0.1 factor after 10 epochs. Finally, to suppress redundant proposals, in all of our experiments we use NMS with $\theta$ = 0.8. We implemented our methods using the Keras framework. Our code is publicly available on \url{https://github.com/hassancpu/Temporal-Action-Localization}. 

\subsection{Comparison with State-of-the-Art Methods}

In this section, we compare our results with state-of-the-art methods. This comparison on Thumos14 is presented in \tablename{ \ref{tab: rm-compar}} according to AR@AN, R@AN = 100-tIoU, and mAP metrics. As seen, our method outperforms previous networks based on these metrics. Specifically, the performance of our method is increased by 1.97\% based on AR@AN=100, and by 6.03\% based on R@AN = 100-tIoU at 0.6 overlap. It is also important to note that the number of parameters of our method, which defines its speed to generate and sort proposals of each video, is 5,546,502, which is, for example, considerably smaller than the RecapNet network \cite{recapnet} with 6,800,000 parameters.

For the sake of a comprehensive analysis, we use the generated proposals on Thumos14 to predict the label of the actions. To this end, following the state-of-the-art works \cite{bsn, tsa}, we use pre-trained UntrimedNet \cite{unet} with top-1 video level label for each proposal. 
As seen, the action detection performance has been improved by 5.12\% based on mAP at IoU@0.7 compared to the best previous results.

On ActivityNet-1.3, we compare our result with state-of-the-art methods based on AR@AN = 100 and AUC metrics in \tablename{ \ref{tab: activity}}. It is evident that our method achieves similar performance to the best of previous methods based on these metrics. It should be noted that, as mentioned before, in order to have a fair comparison with other methods, we used the rescaled features from ActivityNet-1.3 dataset. However, as a result of using these features, we could not benefit from the full potential of our method, e.g., the performance of the interpolation module gets affected by these features.

\subsection{Qualitative Results} 

\begin{figure}[!b]
	\includegraphics[width= \linewidth]{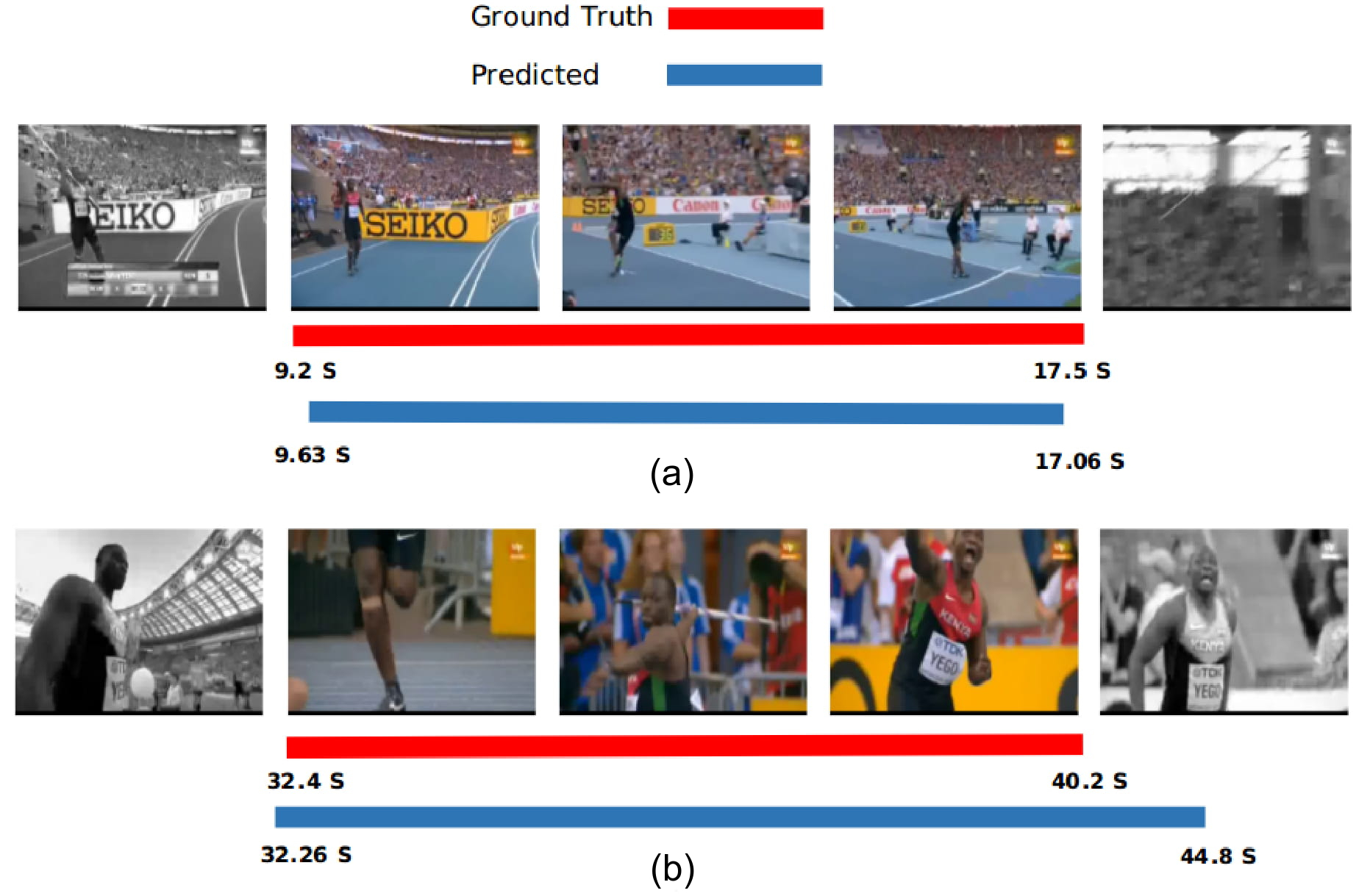}
	\caption{Samples of good generated proposals.}
	\label{fig: prop-great}
	\vspace{1cm}
	\includegraphics[width= \linewidth]{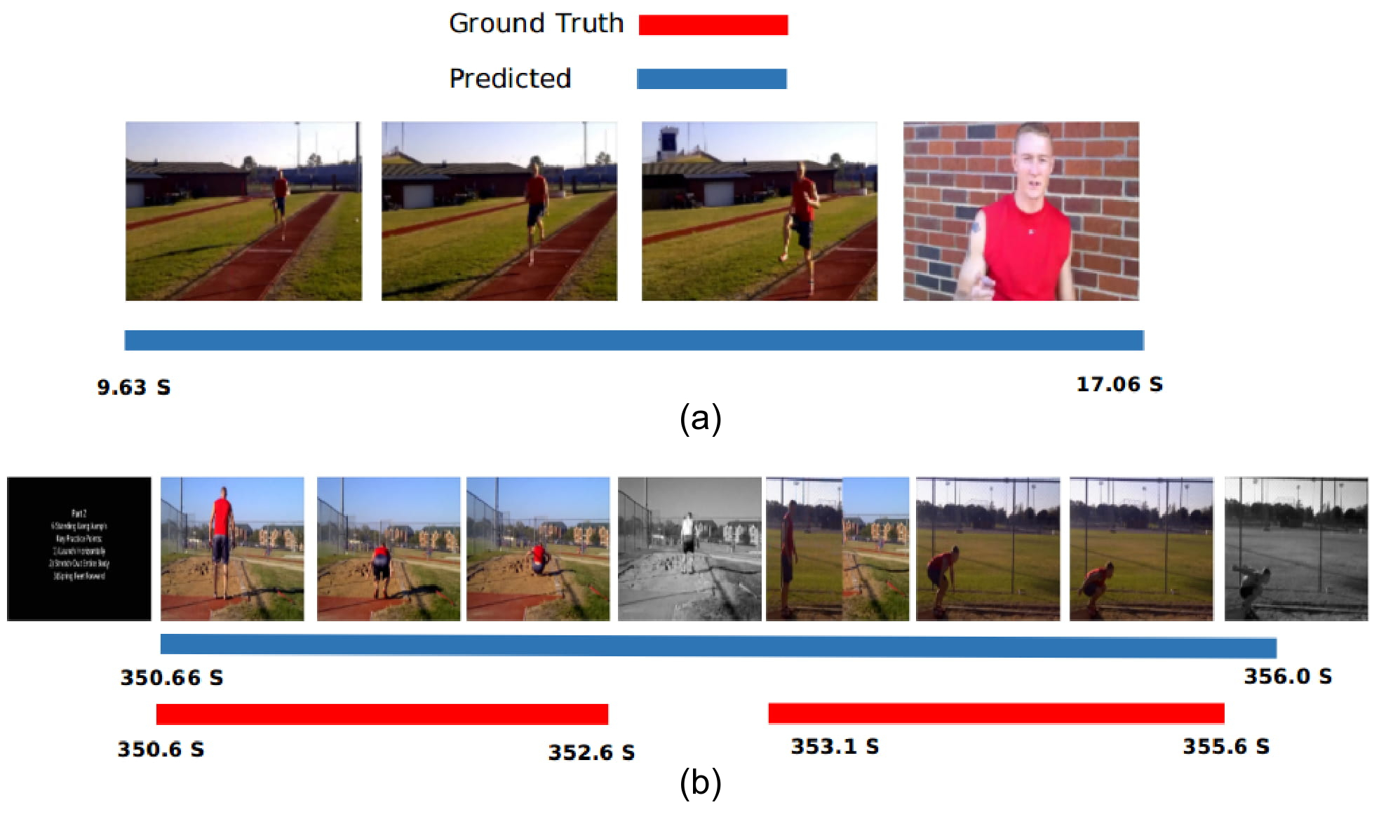}
	\caption{Samples of weak generated proposals.}
	\label{fig: prop-weak}
\end{figure}

Observing some of the generated proposals is useful to identify the pros and cons of the proposed method. So we present some of the good and weak generated proposals. In \figurename{ \ref{fig: prop-great}}(a), the predicted proposal is very accurate and the difference between its temporal boundaries and ground truth is about 0.5s; however in \figurename{ \ref{fig: prop-great}}(b) the end time of the predicted proposal is not accurate (4.6s difference with the ground truth). The reason is that the frames around the end time are so similar to each other which makes it challenging to identify the accurate time of  action's ending. In \figurename{ \ref{fig: prop-weak}}(a), the predicted interval does not correspond to any ground truth proposal, however, we notice that in this interval, one action is happening, but this action is not among the specified actions in the dataset. Finally in \figurename{ \ref{fig: prop-weak}}(b), our network predicts one large proposal, which contains two smaller ground truth proposals, because the interval between those proposals is too small (0.5s).

\begin{table*}[!t]
	\caption{Comparison between different networks performance on the test set based on AR@AN , R@AN=100-tIoU, and mAP metrics on Thumos14}
	\centering
	\renewcommand{\arraystretch}{1.5}
	\begin{tabular}{|c?c|c|c|c|c?c|c|c|c|c?c|c|c|c|c?}
		\hline
		\textbf{Method} & \multicolumn{5}{c?}{\textbf{AR@AN}} &  \multicolumn{5}{c?}{\textbf{R@AN=100-tIoU}} & \multicolumn{5}{c?}{\textbf{mAP}}\\
		\cline{2-16}
		& \textbf{@50} & \textbf{@100} & \textbf{@200} & \textbf{@300} & \textbf{@400} & \textbf{0.5} & \textbf{0.6} & \textbf{0.7} & \textbf{0.8} & \textbf{0.9} & \textbf{0.3} & \textbf{0.4} & \textbf{0.5} & \textbf{0.6} & \textbf{0.7}\\
		\hline
		\textbf{GRU-Split} & 45.15 & 52.34 & 58.64 & 61.80 & 63.72 & \textbf{81.97} & 77.05 & 69.05 & 55.78 & 24.18 & 52.74 & 48.36 & 42.15 & 34.54 & 26.14\\
		\textbf{+ Interpolation} & & & & & &&&&&&&&&&\\
		\textbf{+ Rank Module} &  &  &  &  & &&&&&&&&&&\\
		\hline
		\textbf{GRU-Split} & \textbf{45.53} & \textbf{52.64} & \textbf{59.05} & \textbf{62.32} & \textbf{64.13} & 81.91 & \textbf{77.65} & \textbf{69.48} & \textbf{56.14} & \textbf{24.69} & \textbf{54.62} & \textbf{50.22} & \textbf{44.40} & \textbf{36.40} & \textbf{27.52}\\
		\textbf{+ Interpolation} & & & & & &&&&&&&&&&\\
		\textbf{+ LTR Module} &  &  &  &  & &&&&&&&&&&\\ 
		\hline
	\end{tabular}
	\label{tab: rm-compar-gru}
	\vspace{5pt}
\end{table*}

\section{Ablation Study}

It should be noted that in this ablation study, we first investigate the effect of GRU-split network without the frame interpolation and LTR module. Then we add the frame interpolation module to the GRU-split network to investigate the effect of frame interpolation. Finally, we add LTR module to the other two to investigate the effect of this module on the performance. As a result, for example in case of investigating the effect of GRU-split, instead of frame interpolation, middle frame of the start/end unit is selected as the start/end frame, and instead of LTR module,   score \eqref{eq:1} is used for ranking the proposals.
We first compare the performance of GRU-Split with GRU network. We use the validation set to find the best V and threshold parameters for GRU, which are V = 1 and threshold = 0.3. Other parameters for GRU are the same as for GRU-split, except for the size of the hidden states, which is 513 for GRU-Split to be divisible by three and 512 for GRU network. TAL results for GRU and GRU-Split are reported in \tablename{ \ref{tab: gru-splitted}}. As seen in this table, the GRU-Split outperforms GRU, where for instance, based on AR@AN = 200, 2.72\% improvement and based on R@AN =100-tIoU = 0.5, 4.75\% improvement is achieved. 

Next, we investigate the effect of the interpolation of the frame probabilities on the performance. The results with and without the interpolation method are presented in \tablename{ \ref{tab: frame}}, based on AR@AN and R@AN = 100-tIoU metrics. As shown in the table, the temporal boundaries for the proposals are calculated more accurately using the interpolation method compared to the previous method. 

In order to investigate the effect of LTR module on the performance, we construct a rank module similar to the one used in the start-of-the-art methods \cite{bsn} as follows. The rank module is a FC network with smooth-L1 with  $\delta$ = 0.1 as loss function.
The input of the rank module in \cite{bsn} is 32 proposal features, which are sampled from action probabilities. But here, in order to perform a fair comparison, for both modules, we use the same proposal features. We set the batch size of rank module to 256. More details of the constructed rank module are shown in \figurename{ \ref{fig: rm}}. The comparative results of rank module and LTR module and shown in \tablename{ \ref{tab: rm-compar-gru}}, based on AR@AN, R@AN = 100-tIoU, and mAP metrics. As seen, LTR modules outperforms rank module. Although, we see slight improvements based on AR@AN and R@AN=100-tIoU metrics for the LTR Module compared to the rank module, we observe significant difference between the performance of these two modules based on the mAP metric. Specifically, in IoU@0.5, there is a 2.25\% improvement when the LTR module is used. 

\section{Conclusion}

In this paper, we introduced the GRU-Split network to predict the start, end, and action probabilities of video units. We also proposed an interpolation method to compute the temporal boundaries of the proposals and used a Learn to Rank approach to predict the overlap score of each proposal with the ground truth. Using Thumos14 and ActivityNet-1.3 datasets, we evaluated our proposals and compared the results with state-of-the-art methods which showed the outperformance of our method.	

\section{Compliance with Ethical Standards}
The authors declare that there is no conflict of interest.

\end{document}